\documentclass{article}
\usepackage{spconf,amsmath,graphicx}
\usepackage{times}
\usepackage{epsfig}
\usepackage{amssymb}
\usepackage{color}
\usepackage{multirow}
\usepackage{tabularx}
\usepackage{hyperref}
\usepackage{subcaption}

 \newcommand{\etal}{\textit{et al. }}
 
 \newcommand{\ie}{\textit{i}.\textit{e}. }
\newcolumntype{Y}{>{\centering\arraybackslash}X}

\graphicspath{{./diagrams/}{./figures/fig_qualitative_results/}{./figures/fig_losses/}{./figures/fig_demo/}{./figures/fig_gans_ext/}}

\title{Improving Super Resolution Methods via Incremental Residual Learning}
\name{Muneeb Aadil, Rafia Rahim, Sibt ul Hussain}
\address{ Reveal.ai Lab, National University of Computer and Emerging Sciences, Islamabad, Pakistan\\
\tt \small \{muneeb.aadil, rafia.rahim, sibtul.hussain\}@nu.edu.pk}
\begin{document}
%
\maketitle
\begin{abstract}

Recently, Convolutional Neural Networks (CNNs) have shown promising performance in super-resolution (SR). However, these methods operate primarily on Low Resolution (LR) inputs for memory efficiency but this limits, as we demonstrate, their ability to (i) model high frequency information; and (ii) smoothly translate from LR to High Resolution (HR) space. To this end, we propose a novel Incremental Residual Learning (IRL) framework to address these mentioned issues. In IRL, first we select a typical SR pre-trained network as a master branch. Next we sequentially train and add residual branches to the main branch, where each residual branch is learned to model accumulated residuals of all previous branches. We plug state of the art methods in IRL framework and demonstrate consistent performance improvement on public benchmark datasets to set a new state of the art for SR at only $\approx$ 20\% increase in training time.\footnote{Code available at \url{https://bit.ly/2HipDsJ}}
\end{abstract}


\section{Introduction}
\label{sec:intro}

Single Image Super-Resolution (SR) aims to generate a High Resolution (HR) image $I^{SR}$ from a low
resolution (LR) image $I^{LR}$ such that it is similar to original HR image $I^{HR}$. SR has seen a lot of interest recently because it is: (i) inherently
an ill-posed inverse problem; and (ii) an important low level
vision problem having many applications.

In the wake of CNNs success on vision tasks, many researchers applied them to SR problems. One class of CNN
based methods \cite{dong2016image, kim2016accurate,kim2016deeply,Tai-DRRN-2017,Tai-MemNet-2017} take
interpolated image as input and explicitly model residuals in image space. However, they
have high memory and computational requirements since these networks take an interpolated HR version of an
image as input. To tackle this issue, another class of methods
\cite{Ledig2017PhotoRealisticSI,Lim_2017_CVPR_Workshops,Tong2017ImageSU,lai2017deep,zhang2018residual}
takes $I^{LR}$ as input and apply convolutional operations
primarily in LR space, followed by upsampling layers to produce $I^{SR}$. Although these
networks do not suffer from high memory and computational cost, they have two
design limitations. Firstly, these networks upsample the feature maps too swiftly and thus are unable to reliably
learn the content gap between LR space and HR space. Secondly, these networks cannot explicitly learn residuals in
image space because spatial dimensions of input and output do not match.


\begin{figure}[t]
\centering

  \begin{subfigure}{\linewidth}
    \resizebox{\linewidth}{!}{
    \begin{tabular}{ccc}
      $P_{0}$ & $R_{1}$ & $P_{1} + P_{2}$ (Ours) \\

      \includegraphics[width=0.3\linewidth]{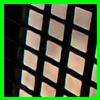} &\includegraphics[width=0.3\linewidth]{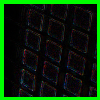} &  \includegraphics[width=0.3\linewidth]{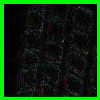} \\
      \includegraphics[width=0.3\linewidth]{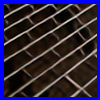} &\includegraphics[width=0.3\linewidth]{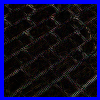} &  \includegraphics[width=0.3\linewidth]{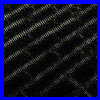} 
    \end{tabular}}
  \end{subfigure}

  \caption{Base network outputs $I^{SR}$ (left) and their residuals (middle) \ie  $I^{HR} - I^{SR}$ are learnt by IRL (right). These residuals are then added to main branch output for improved performance. Please refer to text for further details.}
\end{figure}


We hypothesize that: (i) there is a large content-gap between the LR feature maps and desired HR
output image and this gap cannot be reliably learnt in a single transition step using a
convolutional layer; (ii) instead  of directly modeling HR image, one should explicitly model
residuals -- or High Frequency Information (HFI), in general -- in image space to better model fine level details.

Thus, to this end, we propose the IRL framework: a new learning setup for SR task that improves the
performance of existing networks. IRL uses an existing pre-trained network (\ie master branch)
and sequentially adds other networks (\ie residual branches). Each added residual branch works on
upsampled feature maps, and is trained to learn the accumulated residuals of all the previous branches. At test time, the
output of all branches are added to produce the final output. Our proposed architecture leads to consistent performance
improvement of all the existing state of the art SR networks, adding only 20\% extra training time and no extra memory
overhead because of sequential training of the residuals. 

\section{Related Work}

\begin{figure}[t]
\centering
\includegraphics[width=1\linewidth]{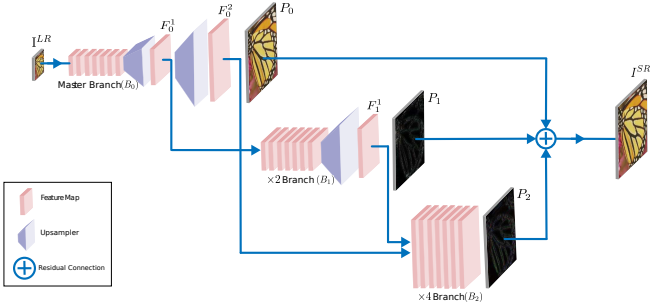}
   \caption{Our proposed IRL framework for $\times4$ scale. Each branch represents an arbitrary image-to-image network. 
   It is evolved from an existing architecture (master branch) by adding the residual branches ($\times2$ and $\times4$).}
\label{fig_irl}
\end{figure}

We decompose our discussion in two separate topics and review them separately in the following
sections.

\subsection{CNNs for Super Resolution} 

SRCNN \cite{dong2016image} first introduced CNNs for learning the
SR task in end-to-end way and showed significant improvement over prior methods. VDSR
\cite{kim2016accurate} then tackled training difficulties encountered in SRCNN \cite{dong2016image}
by explicitly modeling residual learning in image space and proposed very deep 20 layers CNN. Similarly, DRCN \cite{kim2016deeply} introduced  parameters sharing in
network using very deep recursive layers. Afterwards, DRRN \cite{Tai-DRRN-2017}
applied convolutional layers recursively to refine the image iteratively, and MemNet \cite{Tai-MemNet-2017}
employed memory block to take into account hierarchical information. However, since all these networks take as input an interpolated image, residual learning can be employed.
But consequentially, these methods have high memory and computational cost.

To tackle this issue, recent approaches take input $I^{LR}$ and apply convolutional operations
primarily in LR space, followed by upsampling layers to produce $I^{SR}$. Since these methods
apply majority of operations in the LR space, they do not suffer from high memory requirements and
higher computational cost. For instance, SRResNet \cite{Ledig2017PhotoRealisticSI} adapts ResNet
architecture \cite{he2016deep} for SR and then employs ESPCNN \cite{Shi2016RealTimeSI} to upsample
efficiently. Lim \etal \cite{Lim_2017_CVPR_Workshops} improves SRResNet
\cite{Ledig2017PhotoRealisticSI} and propose EDSR baseline (EDSRb) and EDSR. SRDenseNet
\cite{Tong2017ImageSU} employs DenseNet architecture \cite{Huang2017DenselyCC} coupled with skip-connections to
 explicitly model low level features. LapSRN \cite{lai2017deep} employs Laplacian pyramid style
CNNs to iteratively process feature maps and upsample them to refine the earlier prediction. Then, RDN \cite{zhang2018residual} merged residual block of
EDSR \cite{Lim_2017_CVPR_Workshops} and dense skip connections of SRDenseNet \cite{Tong2017ImageSU}
to form Residual Dense Block resulting in even better performance.

\subsection{Objective Functions}\label{sec:review_obj_func}

Earlier networks majorly employ $L_{2}$ loss because: (i) it directly optimizes PSNR, and (ii) it has nice
optimization properties \cite{Wang2009MeanSE}. However, Zhao \etal \cite{Zhao2017LossFF} showed that $L_{1}$ loss outperforms $L_{2}$ in terms of PSNR. Qualitatively speaking, $L_{2}$ recovers
HFI \ie edges better than $L_{1}$ but it leaves splotchy artefacts on plain regions. On the  other end, $L_{1}$ loss removes
splotchy artefacts at the cost of sharper edges recovery. Thus, following \cite{Zhao2017LossFF}, majority of
the networks employ $L_{1}$ loss instead of $L_{2}$ loss.

\section{Proposed Method}\label{sec:method}

In this section, we delineate our proposed framework and training methodology. Firstly,
a SR network is selected as master branch $B_{0}$ and is trained typically on $L_{0} = I^{HR}$. 
When the training converges, its weights are freezed and another network \ie residual branch $B_{1}$ is added. This residual
branch $B_{1}$ takes the upsampled feature maps $F^{1}_{0}$ of master branch $B_{0}$ and is trained on the residuals $R_{1}$ of master branch. This
process is repeated until all scales of feature maps are processed -- \textit{c.f.} figure \ref{fig_irl}. Generally speaking, we can formalize the methodology as follows:


\begin{equation}
P_{i} = \begin{cases}
            B_{i}(I^{LR}) & \text{if } i=0 \\ 
            B_{i}([F_{0}^{i}, F_{1}^{i-1}, \ldots, F_{i-1}^{1}]) & \text{if } i > 0 \\
            
        \end{cases}
\end{equation}

Each branch $B_{i}$ is trained on its respective label $L_{i}$ which is formed as:

\begin{equation}
L_{i} = \begin{cases}
          I^{HR} & \text{if } i=0 \\
          R_{i} & \text{if } i>0
        \end{cases}
\end{equation}

Where $R_{i} = I^{HR} - {\Sigma^{i-1}_{k=0} P_{k}}$. In the above equations, master branch and
residual branches correspond to cases where $i = 0$ and $i > 0$, respectively.

Finally, at test time, result $I^{SR}$ is formed as: 
\begin{equation}
I^{SR} = \Sigma^{n}_{i=0} P_{i}
\end{equation}

Where $n$ is the number of branches and is dependent on SR scale. In our settings, we used a single branch for 
$\times2,\times3$ scales and and 2 branches for  $\times4$ scale.

Following sections explain in details the effects of each design choice of our architecture. 

\subsection{Upsampled Feature Maps} \label{sec:upsampled_feature_maps}

\begin{figure}[h]
\begin{center}
    \begin{subfigure}{0.8\linewidth}
        \includegraphics[width=\linewidth]{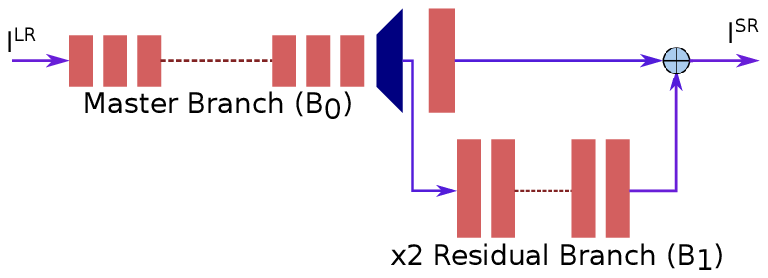}
        \caption{$+^{u}$: $B_{1}$ inputs upsampled feature maps.}
        \label{fig_res_variant1}
    \end{subfigure}

    \begin{subfigure}{0.8\linewidth}
        \includegraphics[width=\linewidth]{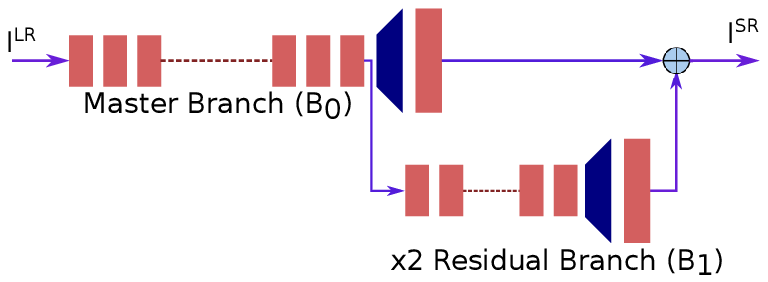}
        \caption{$+^{d}$: $B_{1}$ inputs downsampled feature maps.}
        \label{fig_res_variant2}
    \end{subfigure}    
\end{center}
\caption{Pictorial description of our $+^{u}$ and $+^{d}$ variants.}
\label{fig_res_variants}
\end{figure}

In our setup, each branch operates on higher dimensional feature maps than previous branch. We argue
that this design choice leads to performance enhancement. To confirm this, we trained two variants, as
shown in figure \ref{fig_res_variants}, and measured the performance difference between them. As shown
in table \ref{table_sota_extend}, $+^{u}$ versions lead to enhanced performance
consistently across all networks. Thus, for each branch, we used higher dimensional feature maps
than previous branch. From now onwards, $+/++$ versions take upsampled feature maps, unless specified otherwise.


Now since spatial dimensions are doubled after every branch, each successive branch's parameters have
to be halved also to keep the memory requirements constant. Therefore, we halved number of layers because it empirically showed the best results.


\vspace{-7pt}
\subsection{Residual Learning in Image Space}

Although residual branches can be trained to learn $I^{HR}$ directly, we instead train them on
image residuals because it leads to faster convergence and slightly superior performance \cite{kim2016accurate}.
However, as opposed to \cite{kim2016accurate} which used interpolated image for residuals computation,
we used output of master branch $P_0$ for $B_1$, and sum of $P_0$ and $P_1$ for $B_2$, since earlier
branches are already trained in our scenario.


\vspace{-7pt}
\subsection{Extending Contemporary State of the Arts}

\begin{table}[h]
\begin{center}
\begin{tabular}{ |l|c|c|c|c|c| } 
\hline
 \multirow{2}{*}{Network} & \multirow{2}{*}{Input} & \multirow{2}{*}{Original} & \multicolumn{3}{|c|}{IRL (Ours)} \\
 \cline{4-6}
  &  &  & $+^{d}$ & $+^{u}$ & $++^{u}$ \\
 \hline
 \hline
 VDSR \cite{kim2016accurate} & HR & 30.58 & N/A & 30.58 & 30.58 \\ 
 DRRN \cite{Tai-DRRN-2017} & HR & 30.81 & N/A & 30.81 & 30.81 \\ 
 MemNet \cite{Tai-MemNet-2017} & HR & 30.86 & N/A & 30.86 & 30.86 \\ 
 \hline 
 \hline
 LapSRN \cite{lai2017deep} & LR & 30.67 & 30.69 & \color{blue}{30.73} & \color{red}{30.76} \\
 SRResNet \cite{Ledig2017PhotoRealisticSI} & LR & 29.80 & 29.81 & \color{blue}{29.84} & \color{red}{29.87} \\
 EDSRb \cite{Lim_2017_CVPR_Workshops} & LR & 29.83 & 29.84 & \color{blue}{29.87} & \color{red}{29.89} \\ 
 EDSR \cite{Lim_2017_CVPR_Workshops} & LR & 30.67 & 30.67 & \color{blue}{30.69} & \color{red}{30.71} \\ 
 RDN \cite{zhang2018residual}& LR & 30.64 & 30.65 & \color{blue}{30.67} & \color{red}{30.69} \\
 \hline
\end{tabular}
\end{center}
\caption{Performance in terms of PSNR (dB) of + and ++ versions on validation set for $\times4$ scale.
          Red and blue denotes the best and second best performing methods respectively.}
\label{table_sota_extend}
\end{table}

To isolate the improvement IRL causes to existing state of the arts, we plug contemporary state of the art
networks as our master branch and train only the residual branches from scratch. For uniform comparison, we keep training settings identical to the one used during the training of master branch.


Table \ref{table_sota_extend} shows the effects of adding IRL to the existing methods. IRL consistently
improves the performance of all networks except the ones taking HR image as input. We argue this is because
these networks already employ residual learning in image space and process upsampled
feature maps.

\subsection{Objective Functions} \label{sec:objective_functions}

\begin{figure}[h]
\centering

  \begin{subfigure}{\linewidth}
    \resizebox{.99\linewidth}{!}{
    \begin{tabular}{ccc}
      HR (PSNR) & $L_{1}$ (23.19 dB) & $L_{2}$ (\textbf{23.43} dB) \\
      \includegraphics[width=0.3\linewidth]{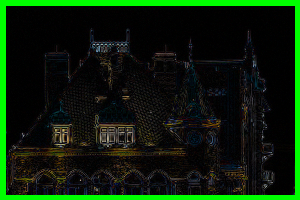} &\includegraphics[width=0.3\linewidth]{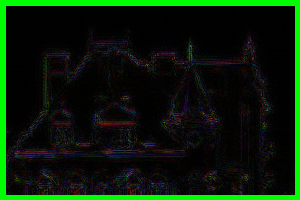} &  \includegraphics[width=0.3\linewidth]{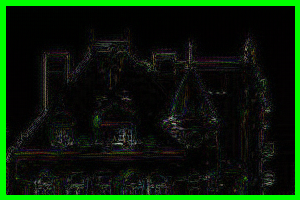} \\
      HR (PSNR) & $L_{1}$ (24.13 dB) & $L_{2}$ (\textbf{24.22} dB) \\
      \includegraphics[width=0.3\linewidth]{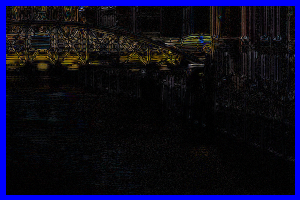} &\includegraphics[width=0.3\linewidth]{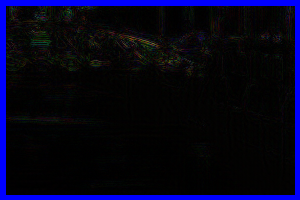} &  \includegraphics[width=0.3\linewidth]{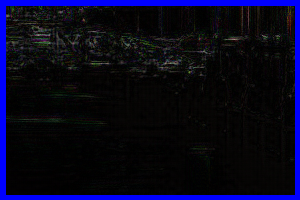} 
    \end{tabular}}
  \end{subfigure}

\caption{Residuals $R_{1}$ (left), learning capability of $L_{1}$ loss (middle), and of $L_{2}$ loss (right). $L_{2}$ performs better than $L_{1}$ for capturing details of residuals.}
\label{fig_losses}
\end{figure}


Existing methods have a trade-off between shaper edges and higher PSNR since it can be trained
with only one loss function at a time due to having a single branch. In comparison since our
framework has multiple branches, we can use different objective functions at the same time
for modeling different information by each branch.

Specifically, different from the models trained in previous section, we trained the residual
branches with $L_{2}$  loss instead of $L_{1}$ loss. Figure \ref{fig_losses} shows the
comparison: $L_{2}$ loss not only recovers shaper edges, but it also performs better in terms
of PSNR as opposed to previous works \cite{Lim_2017_CVPR_Workshops,Zhao2017LossFF}. We argue this is
because previous works model HR image directly. On the other hand, we explicitly model HFI. 


\begin{table*}[ht]
\begin{center}
\begin{tabularx}{\textwidth}{|c|l|Y|Y|Y|Y| }
\hline 
Scale & Method & Set5 \cite{bevilacqua2012low} & Set14 \cite{zeyde2010single} & B100 \cite{martin2001database} & Urban100 \cite{huang2015single} \\
\hline
\hline
\multirow{8}{*}{$\times2$} 
& LapSRN \cite{lai2017deep} & 37.52 / 0.9573 & 33.08 / 0.9121 & 31.80 / 0.8924 & 30.41 / 0.9091 \\
& LapSRN+ (Ours) & \textbf{37.55} / \textbf{0.9586} & \textbf{33.11} / \textbf{0.9122} & \textbf{31.82} / \textbf{0.8927} & \textbf{30.45} / \textbf{0.9093} \\
\cline{2-6}
& EDSRb \cite{Lim_2017_CVPR_Workshops} & 37.98 / 0.9604 & 33.56 / 0.9173 & 32.15 / 0.8994 & 31.97 / 0.9271 \\
& EDSRb+ (Ours) & \textbf{38.01} / \textbf{0.9606} & \textbf{33.58} / \textbf{0.9174} & \textbf{32.18} / \textbf{0.8997} & \textbf{32.03} / \textbf{0.9276} \\
\cline{2-6}
& EDSR \cite{Lim_2017_CVPR_Workshops} & 38.11 / 0.9602 & 33.92 / 0.9195 & 32.32 / 0.9013 & 32.93 / 0.9351 \\
& EDSR+ (Ours) & \textbf{38.14} / \textbf{0.9604} & \textbf{33.94} / \textbf{0.9196} & \textbf{32.34} / \textbf{0.9014} & \textbf{32.96} / \textbf{0.9354} \\
\cline{2-6}
& RDN \cite{zhang2018residual} & 38.23 / 0.9613 & 34.00 / 0.9211 & 32.33 / 0.9016 & 32.87 / 0.9352 \\
& RDN+ (Ours) & \textbf{38.27} / \textbf{0.9615} & \textbf{34.03} / \textbf{0.9214} & \textbf{32.36} / \textbf{0.9019} & \textbf{32.91} / \textbf{0.9355} \\

\hline 
\hline

\multirow{6}{*}{$\times3$} 
& EDSRb \cite{Lim_2017_CVPR_Workshops} & 34.36 / 0.9267 & 30.28 / 0.8415 & 29.08 / 0.8053 & 28.14 / 0.8525 \\
& EDSRb+ (Ours) & \textbf{34.41} / \textbf{0.9271} & \textbf{30.31} / \textbf{0.8417} & \textbf{29.11} / \textbf{0.8056} & \textbf{28.17} / \textbf{0.8528} \\
\cline{2-6}
& EDSR \cite{Lim_2017_CVPR_Workshops} & 34.65 / 0.9282 & 30.52 / 0.8462 & 29.25 / 0.8093 & 28.80 / 0.8653 \\
& EDSR+ (Ours) & \textbf{34.68} / \textbf{0.9284} & \textbf{30.55} / \textbf{0.8465} & \textbf{29.28} / \textbf{0.8096} & \textbf{28.83} / \textbf{0.8655} \\
\cline{2-6}
& RDN \cite{zhang2018residual} & 34.70 / 0.9293 & 30.56 / 0.8467 & 29.25 / 0.8094 & 28.79 / 0.8651 \\
& RDN+ (Ours) & \textbf{34.73} / \textbf{0.9294} & \textbf{30.60} / \textbf{0.8469} & \textbf{29.27} / \textbf{0.8096} & \textbf{28.81} / \textbf{0.8654} \\

\hline 
\hline 

\multirow{15}{*}{$\times4$} 
& LapSRN \cite{lai2017deep} & 31.62 / 0.8869 & 28.12 / 0.7718 & 27.33 / 0.7286 & 25.28 / 0.7602 \\
& LapSRN+ (Ours) & 31.64 / 0.8867 & 28.16 / 0.7716 & 27.35 / 0.7289 & 25.33 / 0.7608 \\
& LapSRN++ (Ours) & \textbf{31.67} / \textbf{0.8870} & \textbf{28.18} / \textbf{0.7718} & \textbf{27.37} / \textbf{0.7290} & \textbf{25.36} / \textbf{0.7612} \\
\cline{2-6}
& SRResNet \cite{Ledig2017PhotoRealisticSI} & 32.05 / 0.8910 & 28.53 / 0.7804 & 27.57 / 0.7354 & 26.07 / 0.7839 \\
& SRResNet+ (Ours) & 32.08 / 0.8913 & 28.55 / 0.7807 & 27.59 / 0.7356 & 26.13 / 0.7844 \\
& SRResNet++ (Ours) & \textbf{32.10} / \textbf{0.8914} & \textbf{28.57} / \textbf{0.7808} & \textbf{27.62} / \textbf{0.7358} & \textbf{26.15} / \textbf{0.7845} \\
\cline{2-6}
& EDSRb \cite{Lim_2017_CVPR_Workshops} & 32.09 / 0.8926 & 28.56 / 0.7808 & 27.56 / 0.7359 & 26.03 / 0.7846 \\
& EDSRb+ (Ours) & 32.13 / 0.8932 & 28.59 / 0.7816 & 27.59 / 0.7362 & 26.07 / 0.7848 \\
& EDSRb++ (Ours) & \textbf{32.15} / \textbf{0.8934} & \textbf{28.60} / \textbf{0.7817} & \textbf{27.60} / \textbf{0.7362} & \textbf{26.08} / \textbf{0.7849} \\
\cline{2-6}
& EDSR \cite{Lim_2017_CVPR_Workshops} & 32.46 / 0.8968 & 28.80 / 0.7876 & 27.72 / 0.7420 & 26.64 / 0.8033 \\
& EDSR+ (Ours) & 32.48 / 0.8970 & 28.83 / 0.7878 & 27.75 / 0.7423 & 26.66 / 0.8034 \\
& EDSR++ (Ours) & \textbf{32.49} / \textbf{0.8971} & \textbf{28.84} / \textbf{0.7879} & \textbf{27.75} / \textbf{0.7424} & \textbf{26.69} / \textbf{0.8036} \\
\cline{2-6}
& RDN \cite{zhang2018residual} & 32.46 / 0.8980 & 28.80 / 0.7868 & 27.71 / 0.7420 & 26.61 / 0.8027 \\
& RDN+ (Ours) & 32.48 / 0.8982 & 28.82 / 0.7871 & 27.74 / 0.7424 & 26.63 / 0.8030 \\
& RDN++ (Ours) & \textbf{32.50} / \textbf{0.8982} & \textbf{28.84} / \textbf{0.7872} & \textbf{27.75} / \textbf{0.7424} & \textbf{26.64} / \textbf{0.8031} \\
\hline

\end{tabularx}
\end{center}
   \caption{Quantitative Results (PSNR (dB) / SSIM) of adding IRL framework to the existing state of the art
   methods on different benchmarks datasets. Our proposed framework leads to
consistent performance improvement for all scales.}
\label{table_quantitative_results}
\end{table*}
\section{Experiments and Results}
We trained our network on DIV2K dataset \cite{Agustsson_2017_CVPR_Workshops} which contains 800 training images. We used first 790 images for training and the remaining 10 images for validation set. For testing, we used following benchmark datasets: Set5 \cite{bevilacqua2012low}, Set14 \cite{zeyde2010single}, B100 \cite{martin2001database}, and Urban100 \cite{huang2015single}.

For the training of residual branches, we used ADAM \cite{Kingma2014AdamAM} optimizer with learning rate set to $10^{-4}$ and $\beta_{1} = 0.9, \beta_{2} = 0.99$. We set input LR image patch size according to the one used in the original master branch to ensure uniform comparison. For more details and qualitative results, please refer to our publicly available code. For quantitative comparison, we evaluated our approach in terms of PSNR and SSIM on Y channel against the following state of the art networks: LapSRN\cite{lai2017deep}, SRResNet  \cite{Ledig2017PhotoRealisticSI}, EDSRb
\cite{Lim_2017_CVPR_Workshops}, EDSR \cite{Lim_2017_CVPR_Workshops}, RDN \cite{zhang2018residual} -- \textit{c.f.} Table \ref{table_quantitative_results}. We skipped (i) SRDenseNet because of unavailability of pre-trained weights and public implementation of the method, and (ii) networks taking interpolated image as input because it already processes upsampled feature maps and learn residuals in image space.
\section{Conclusions}\label{sec:conclusions}
Firstly, we argued that processing majorly in LR space is sub-optimal because (i) transition to HR space
in single convolutional layer is unreliable, and (ii) it is not possible to learn residuals in image space which
leads to performance loss. Secondly, we empirically showed that for learning residuals, $L_{2}$ loss
performs better than $L_{1}$ loss -- unlike previous research -- as it better models the HFI.

We addressed aforementioned issues by introducing residual branches which operate on upsampled feature
maps, and learn residuals in image space through $L_{2}$ loss. Our proposed framework leads to consistent
improvement and sets a new state of the art for SR at marginal training time cost.
\section{Acknowledgments}
We would like to acknowledge the computational grant from Higher Education Commission (HEC) of Pakistan.
\bibliographystyle{IEEEbib}
\bibliography{strings,refs,egbib}

\end{document}